\documentclass[10pt,twocolumn,letterpaper]{article}

\usepackage{wacv}
\usepackage{times}
\usepackage{epsfig}
\usepackage{graphicx}
\usepackage{amsmath}
\usepackage{amssymb}
\usepackage{multirow}
\wacvfinalcopy % *** Uncomment this line for the final submission

\def\wacvPaperID{12} % *** Enter the wacv Paper ID here

\ifwacvfinal\fi
\begin{document}

%%%%%%%%% TITLE
\title{Syn2Real: Forgery Classification via Unsupervised Domain Adaptation}

% Authors at the same institution
\author{Akash Kumar \hspace{2cm} Arnav Bhavasar \\
IIT Mandi\\
{\tt\small akash\_bt2k15@dtu.ac.in, arnav@iitmandi.ac.in}
}
% Authors at different institutions
% \author{First Author \\
% Institution1\\
% {\tt\small firstauthor@i1.org}
% \and
% Second Author \\
% Institution2\\
% {\tt\small secondauthor@i2.org}
% }

\maketitle
\ifwacvfinal\thispagestyle{empty}\fi

%%%%%%%%% ABSTRACT
\begin{abstract}
   In recent years, image manipulation is becoming increasingly more accessible, yielding more natural-looking images, owing to the modern tools in image processing and computer vision techniques. The task of the identification of forged images has become very challenging. Amongst different types of forgeries, the cases of Copy-Move forgery are increasing manifold, due to the difficulties involved to detect this tampering. To tackle such problems, publicly available datasets are insufficient. In this paper, we propose to create a synthetic forged dataset using deep semantic image inpainting and copy-move forgery algorithm. However, models trained on these datasets have a significant drop in performance when tested on more realistic data. To alleviate this problem, we use unsupervised domain adaptation networks to detect copy-move forgery in new domains by mapping the feature space from our synthetically generated dataset. Furthermore, we improvised the F1 score on CASIA and CoMoFoD dataset to 80.3\% and 78.8\%, respectively. Our approach can be helpful in those cases where the classification of data is unavailable.
\end{abstract}

%%%%%%%%% BODY TEXT
\section{Introduction}

With the advancement of new image editing technologies, there is a sharp increase in the  number of forgery cases. While sophisticated image editing tools are meant to enhance the quality of images, they are misused to create forged images for nefarious purposes. These images look so natural that it is difficult to tell with naked eyes whether they have been tampered or are they authentic. It has led to a rise in the cases of image forgery in several fields - medical imaging, industrial photography, surveillance system, criminal, and forensic investigation. \cite{security} 

There are diverse ways of forging images, of which Copy-Move, Splicing, Retouching, and Resampling forgeries are the most common ones. Copy-Move Forgery (CMF) is a type of passive image forgery technique in which a section of an image is copied and pasted within the same image. Many post-image processing operations such as rescaling, affine transformations, resizing, and blurring are applied to the copied region. As the source and target image remains the same, the photometric characteristics of the image remain mostly invariable. Thus, the detection becomes even more difficult. For instance, in contrast to CMF, splicing forgery is a composition of two images. A section is cut from an image and pasted on another image. As a result, there is an edge discrepancy that makes the detection of splicing forgery relatively easier.

% \textbf{Motivation} 
Image tampering can have significant effects in various domains. For instance, in medical imaging, the images are procured with the utmost sensitivity and is a tiresome process. There can be ulterior economical motives for tampering these confidential and sophisticated images. Consequently, it could misguide the patients about their illnesses and injuries. In the field of education, students can tamper their documents with online available software tools. The significant impact of image tampering can happen in the socio-political area, as manipulated images can affect the perception of a large group of people. Many magazines and newspaper editors tamper the images in such a way that they can change the semantic meaning of the image.

There have been several traditional approaches for forgery detection that include mostly block-based and keypoint feature extraction \cite{lit_1, lit_2, lit_3} and matching procedures. Nowadays, deep learning approaches \cite{Zhang2016ImageRF, augment, busternet} have been proposed to counterattack the problem of image forgery. However, most of the approaches are based on supervised learning. When there are a lot of labeled examples, then it is easy to train the model via supervised learning. To counter the problems of training data, we generally surrogate the training data by including the dataset from adjacent modality or use synthetic imagery. When the same model is evaluated on these datasets, it results in a significant drop in the performance. It happens due to the shift in style, content, or appearance distribution between various datasets. 
In these cases, domain adaptation is needed to learn the distribution shift.
% For example, there are diverse ways to create synthetic images. Finetuning models on these datasets may perform well during the test time on these synthetic datasets, but there is a drop in performance when the same model is tested in other real-world scenarios.

% Image Region Forgery Detection \cite{Zhang2016ImageRF}
\begin{figure*}[t]
\centering
\includegraphics[width=0.9\linewidth]{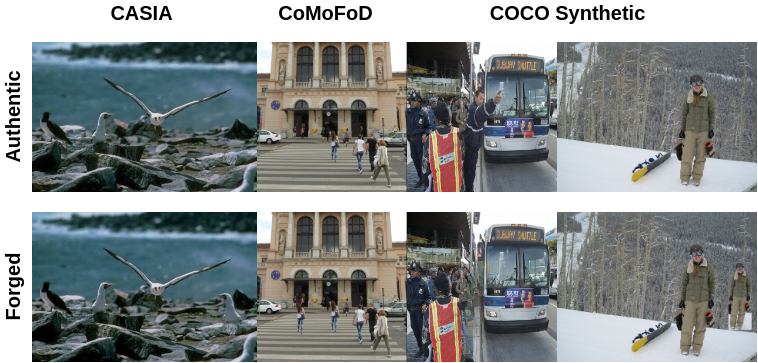} % Reduce the figure size so that it is slightly narrower than the column. Don't use precise values for figure width.This setup will avoid overfull boxes. 
\caption{The first and second column shows the example of target domain dataset (CASIA and CoMoFoD respectively). Subsequent column shows the generated synthetic data from COCO dataset using semantic inpainting and copy-move forgery algorithm. First row is authentic image of each category and second row is forged image.}
\label{fig3}
\end{figure*}

In this work, we show that manipulations in images across different domains can be detected via domain adaptation. We leverage the power of Convolutional Neural networks (CNNs) to perceive the distinguishable features of authentic and tampered images. We tackle the problem of performance drop by incorporating the feature space alignment between our synthetic generated datasets and datasets that are publicly available. We generate the synthetic dataset using Edge-connect semantic inpainting and CMF algorithm. 

\textbf{Contributions} Our main contributions in the paper are summarized as follows: 1) The primary task is to classify images as forged or authentic, for which we employ Unsupervised Domain Adaptation (DA), due to the difference in content and style between our source and target dataset, 2) As the publicly available datasets are small, we generate a new dataset comprising of 80,000 images using deep semantic inpainting and copy-move forgery algorithms on COCO \cite{coco} dataset, and 3) We explore two Unsupervised DA methods to adapt the features from source dataset to target dataset, such that the variation between the domains is minimized. 

The paper is organized as follows: Section II describes the traditional and deep learning solutions that evolved over the years for forgery classification and a review of domain adaptation methods. Section III describes our methodology in detail that involves dataset generation, Unsupervised DA, and final architectures used for training. After that, in Section IV, we evaluate the performance of our architecture on CASIA\cite{casia} and CoMoFoD\cite{comofod} dataset. Section V discusses the conclusion and future directions of our work.

%-------------------------------------------------------------------------
\section{Related Work}

Early works on forgery detection involved traditional approaches such as Block-based and Keypoint-based. In recent years, with the emerging applications of neural nets, more work is motivated using CNNs. Considering issues in the forgery detection areas, like small datasets or unavailability of labels, now there is work that has started focusing on unsupervised learning. Below, we briefly describe some earlier and contemporary work.

% DCT, DWT, DyWt, and FWHT - FT methods
Block-based approaches divide the image into chunks of the block that may overlap each other. Then, features are extracted from these blocks and matched against each other to compare the similarity between blocks. For feature extraction, the primary methods used are Frequency transform, Texture and intensity-based features, moments, invariant features, and dimensional reduction techniques. Frequency transform methods \cite{lit_2} are robust to JPEG compression and helps reduce computational complexity, albeit, the main disadvantage was that they were only restricted to JPEG compressed images. Their performance was not checked on TIFF image formats.  Texture and Identity-based features were incapable to deal with variable angle rotation in images. Moment invariant methods such as Central and Zernike are robust to translation, scaling, rotation, and contrast changes with improvement in time. In dimensional reduction techniques, there is a loss in image details and, thus, low performance in JPEG compressed images. Coming to the block-based matching, Lexicographical Matching, Hashing, Euclidean Distance are some of the approaches used to match the blocks based on defined thresholds. However, much unwanted time is wasted in comparing the similarity between these blocks.

In Key-point methods, SIFT and SURF are primarily used to extract features from images. The main shortcomings are the failure to recognize feature points in flat areas and detect forged regions if the copy and pasted area looks naturally similar. For example, an area of grass is cropped and pasted over the same region where the grass is present. Then, no helpful features can be extracted for matching purposes. In the keypoint feature matching process, we need to keep a check on several hyperparameters involving similarity matching threshold and window size for comparison of patches. They also encounter high time complexity as they have to match so many number of feature points. 

More recently, CNNs based approaches have started evolving with the ability of neural networks to memorize complex visual features. This can be contemplated by the admirable performance of CNNs in various domains.  Zhang \etal \cite{Zhang2016ImageRF} divides images into 32x32 patches and matches them using nearest neighbourhood algorithm. In \cite{dl_2}, forgeries are detected using hand-crafted filters and features. This steganalysis approach helps to identify diverse types of forgeries, but, there are hundreds of parameters that need to be tuned manually. \cite{dl_3} applies self-correlation between different blocks extracted from an image. Then, they employ a pointwise feature extractor to match similar points and generates a mask over a forged region. Busternet,\cite{busternet} the further work of this paper, uses a two branch architecture, to localize the manipulated regions. They created a dataset of one lakh images and then applied a supervised pixel label classification to learn forged features.

\begin{figure*}[htbp]
\centering
\includegraphics[width=0.8\textwidth]{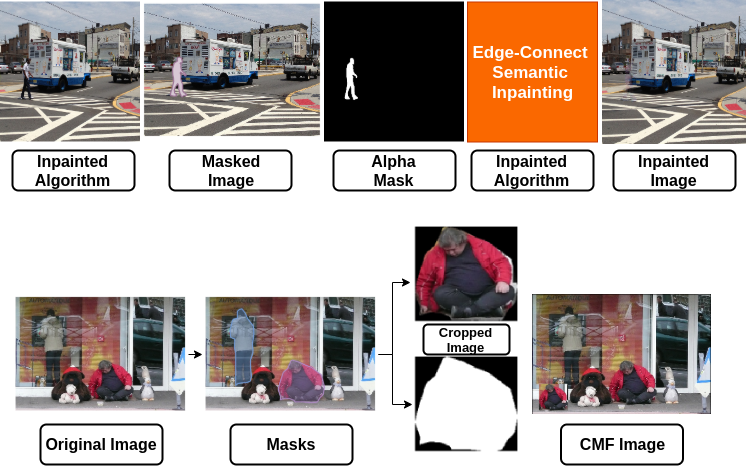}
\caption{Dataset Generation method: In first approach, Edge semantic inpaint fills the region where object is removed. In CMF, a specific category is picked and copy-pasted on the same image. We generated a dataset of ~80,000 images.}
\label{fig2}
\end{figure*}

In DA, initially, statistical-based approaches were used to minimise the distribution shift such as Correlation Alignment, Maximum Mean Discrepancy (MMD) and Kullback-Leibler divergence. \cite{uda_survey} They calculate the distance and apply linear transformation to align the features of source domain to target domain before sending them for classification. The latest advances using CNNs helps to learn more transferable features. Domain Adverarial Neural Network (DANN), Deep Domain Confusion (DDC), Adversarial Discriminative Domain Adaptation and Joint Adaptation Network \cite{uda_survey} are few of the methods that have illustrated a significant improve in unsupervised DA. DANN and DDC use a shared feature extraction layer and then apply minimax and domain confusion loss to adapt the feature space. Annadani's \etal work is the first use case of domain adaptation in the field of image forgery. They used supervised domain adaptation using the Maximum Mean Discrepancy (MMD) loss criterion to decrease the domain discrepancy. Our work is built upon the one reported in \cite{augment} and \cite{dann}. The synthetic dataset created by them only included a feathering approach to create an image, which is more suitable for spliced forgery cases. It does not adapt very well if the image is cut-pasted on the same image, as the characteristics as a whole remain consistent.

\section{Methodology}

\subsection{Dataset Generation}

We applied two methods to generate the dataset.  The inclusion of any one of them shows an increase in performance. Semantic Inpainting helps the model to learn edge discrepancies when the objects are removed. Copy-Move tampered images improve the focus of the network to recognize similar patches.

\subsubsection{Semantic Inpainting}
We synthesized a dataset of inpainted images using all the sub-categories of the COCO dataset \cite{coco} equally. The mask of particular sub-categories was cropped out. After that, Edge-connect \cite{edgeconnect} inpainting is used. It is a type of Deep Semantic inpainting that uses a two-stage approach to complete an image. Firstly, the edge generator fills out the missing edges, and then the image completion network completes the image based on the edges deduced.  Using the above approach, we created approximately 20,000 inpainted images.

 \subsubsection{Copy-Move Forgery: Using Object ROI}
To generate copy-paste forged images, we selected a specific category of the COCO dataset. Then, we took into consideration of each mask's area belonging to that category. Comparing all the mask areas, we select the mask with the largest area. Now, this copied area is pasted over the image after affine transformations and image blending operations. Blending helps the image to fuse in another image smoothly. If an object is just copy-paste, then there is a distinctive boundary that can easily be detected by the algorithm. It's not helpful at the test time, as, in both CASIA and CoMoFoD dataset, the copied areas are post-processed and then pasted. Henceforth, we applied a combination of alpha blending (using eqn \eqref{blend}) and Deep Image Matting\cite{image_matting} operations so that the pasted regions could easily fit on the second image. With this approach, we created a dataset of approximately 60,000 images.
Figure \ref{fig2} illustrates the method of dataset generation.

\begin{equation}
    I \textsubscript{\textit{f}} = \alpha F + (1 - \alpha)B \label{blend}
\end{equation}
where I\textsubscript{\textit{f}} is final image, F is foreground object, B is background image and $\alpha$ is the blending factor.

\subsection{Unsupervised Domain Adaptation}

\begin{figure}[htbp]
\centering
\includegraphics[width=0.9\columnwidth]{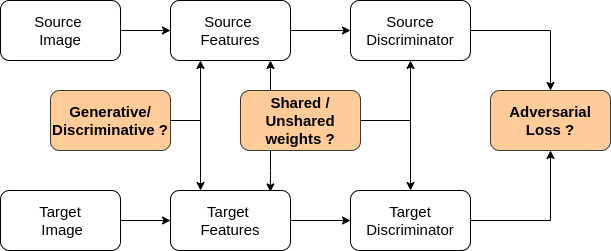} % Reduce the figure size so that it is slightly narrower than the column. Don't use precise values for figure width.This setup will avoid overfull boxes. 
\caption{Generalized architecture of Unsupervised Domain Adaptation algorithms}
\label{fig4}
\end{figure}

Domain Adaptation (DA) is useful in the areas where the dataset with labels are present in a small amount. Usually, most of the domain adaptation task focuses on an unsupervised approach as the number of target labels present is very small in real-world and labels are mostly absent. Unsupervised DA has a generalized architecture that depends on three parameters. Firstly, it depends on the base model, whether it is generative or discriminative, next, based on weight sharing between source and target domain, and, lastly what type of adversarial loss is used (see Fig. \ref{fig4}). In our paper, we explored two unsupervised DA methods, that are DANN\cite{dann} and DDC\cite{ddc}. Both of them have a discriminative base model, and the weights between the layers are shared. The adversarial loss in the case of DANN is minimax, whereas, in DDC, it's confusion loss. The detailed architectures are discussed in the subsequent subsections.

\subsubsection{DANN}

This approach involves the creation of a feature space of images present in the source and target domain. We aim to create a distribution of feature representations such that it is discriminatory among classes and invariant across domains. At the training time, images of both the domains are present, but we have access to class labels of only the source domain. DANN has two separate heads, Source Classification (SC) head, and Domain Classification (DC) head. In SC, feature mapping parameters($\theta$\textsubscript{\textit{f}}) and label classifier parameters($\theta$\textsubscript{\textit{s}}) are optimized as such to reduce the classification loss(\textit{L\textsubscript{s}}) in case of the source domain \eqref{eq_1}. While in DC, feature mapping parameters($\theta$\textsubscript{\textit{f}}) maximizes the domain loss(\textit{L\textsubscript{d}}) so that the distribution of both domains becomes similar \eqref{eq_2}. It simultaneously minimizes the classification loss for the image, whether it comes from a source or target distribution. Let source ,feature and target distributions be denoted as D\textsubscript{s}(x,y), D\textsubscript{f}(x,y) and D\textsubscript{t}(x,y), respectively, where x is input image feature space and y is the label annotated to that image.  
\begin{equation}
    (\hat{\theta} \textsubscript{\textit{f}}, \hat{\theta} \textsubscript{\textit{s}}) = arg min L(\theta \textsubscript{\textit{f}}, \theta \textsubscript{\textit{s}}, \theta \textsubscript{\textit{d}})\label{eq_1}
\end{equation}
\begin{equation}
    \hat{\theta} \textsubscript{\textit{d}} = arg max L(\theta \textsubscript{\textit{f}}, \theta \textsubscript{\textit{s}}, \theta \textsubscript{\textit{d}})
    \label{eq_2}
\end{equation}

\begin{figure}[htbp]
\centering
\includegraphics[width=0.8\columnwidth]{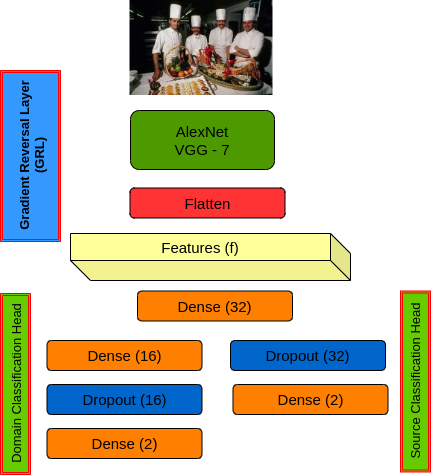} % Reduce the figure size so that it is slightly narrower than the column. Don't use precise values for figure width.This setup will avoid overfull boxes. 
\caption{DANN model is divided into three components: i) feature extractor, ii) source classifier, and iii) domain classifier. Joint loss is used to improve the network's ability to assign category and domain labels to images.}
\label{fig1}
\end{figure}

In this way, the network increases the confusion between source and target domain, so that the model focuses more on the features that help to distinguish images amongst different classes. As mentioned above, DANN optimizes two loss functions concurrently during backpropagation. To implement the losses, Stochastic gradient descent can't be directly applied. So, the authors introduced a particular layer called Gradient Reversal Layer (GRL), which is present in the DC network. During forward pass, it acts as an identity transform. At the time of backpropagation, it multiplies the gradient by a negative constant (-$\Delta$) and passes it to the previous layers. The overall loss function is defined in the \eqref{dann_loss}. 

\begin{eqnarray}
    L(\theta \textsubscript{\textit{f}}, \theta \textsubscript{\textit{s}}, 
    \theta \textsubscript{\textit{d}})&=&\sum_{x=1...N} L \textsubscript{s} (D \textsubscript{s}(D\textsubscript{f}), y\textsubscript{i}) \nonumber \\
    && + (\Delta) \sum_{x=1...N} L \textsubscript{d} (D\textsubscript{t}(D\textsubscript{f}), y\textsubscript{i})
    \label{dann_loss}
\end{eqnarray}{}

\subsubsection{DDC}

Using domain confusion loss, DDC learns the mapping of the source domain. It minimizes the distance between the source and target distributions via Maximum Mean Discrepancy (MMD) loss. The architecture separately learns the discriminative features needed to classify via supervised learning using source images and labels and features required to classify the domain of the image. Let the mapping of source domain be D\textsubscript{s} and mapping of target domain be D\textsubscript{t}, then the network aims to learn a representation that could easily be transferable across various domains. The joint loss function used to train this architecture comprises of categorical cross entropy loss (\textit{L\textsubscript{classify}}) and penalty parameter($\alpha$) multiplied by MMD loss.

\begin{equation}
    L \textsubscript{total} = L \textsubscript{\textit{classify}} (x, y) + \alpha MMD\textsuperscript{2}(D\textsubscript{s}, D\textsubscript{t})
    % argmin \sum{x=1}^{X} log C(M \textsubscript{\textit{s}}(x \textsubscript{\textit{s}})
\end{equation}
where  MMD (D\textsubscript{s}, D\textsubscript{t}) is the distance between the domains calculated using eqn.\eqref{mmd}, and, d\textsubscript{s} and d\textsubscript{t} are data points from source and target domains respectively.

\begin{eqnarray}
    MMD(D\textsubscript{s}, D\textsubscript{t}) = \left \| \frac{1}{\left | D\textsubscript{s} \right |} \sum_{d\textsubscript{s} \epsilon D\textsubscript{s}} \phi \left ( d\textsubscript{s} \right ) - \frac{1}{\left | D\textsubscript{t} \right |} \sum_{d\textsubscript{t} \epsilon D\textsubscript{t}} \phi \left ( d\textsubscript{t} \right ) \right \|\label{mmd}
\end{eqnarray}{}

% \begin{equation}
% \resizebox{.9\hsize}{!}{$ MMD(D\textsubscript{s}, D\textsubscript{t}) = \left \| \frac{1}{\left | D\textsubscript{s} \right |} \sum_{d\textsubscript{s} \epsilon D\textsubscript{s}} \phi \left ( d\textsubscript{s} \right ) - \frac{1}{\left | D\textsubscript{t} \right |} \sum_{d\textsubscript{t} \epsilon D\textsubscript{t}} \phi \left ( d\textsubscript{t} \right ) \right \|$}
% \end{equation}

The main difference compared to DANN is the use of confusion loss. In DDC, domain confusion and domain classifier loss are separate. Domain confusion maximizes the loss across domains, whereas, in DANN, minimax loss, simultaneously converges two loss functions. In DANN and DDC, the weights are tied, making it symmetric mapping between the domains. It enforces both the domains to have the same representations in feature space.

\begin{figure}[htbp]
\centering
\includegraphics[width=\columnwidth]{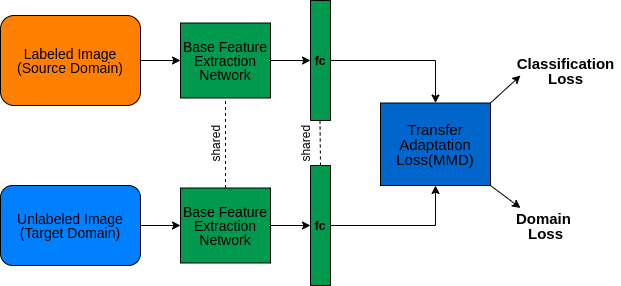} % Reduce the figure size so that it is slightly narrower than the column. Don't use precise values for figure width.This setup will avoid overfull boxes. 
\caption{The DDC network use a shared feature extraction layer to adapt the features of target domain. Meanwhile, transfer adaptation loss minimizes the distribution shift between the \textit{fc} layers.}
\label{ddc}
\end{figure}

\subsection{Architecture}

The base model of our architecture extracts features from images. In DANN, after feature extraction, the DC network predicts the domain of the input image, and the SC network predicts the label for source samples. Figure \ref{fig1} depicts the architecture of our proposed approach. 
At the time of training, we know whether the source domain image is authentic or forged, while we did not use labels of the target domain. We used binary labels to indicate whether the input image comes from the source or target distribution. In the case of DDC, the last \textit{fc} layer features compute the MMD distance between the domains. This distance is then backpropagated to minimize the classification loss and maximize the domain confusion loss. 

\section{Experiments Analysis}

\subsection{Dataset Tested}
We evaluated our architecture on CASIA V2 and CoMoFoD datasets. In our case, the source domain constitutes of COCO CMF and semantic inpainted images, and, target domain comprises CASIA V2 and CoMoFoD datasets. Exhaustive experiments were done using AlexNet \cite{alexnet} and VGG-7 \cite{vgg}  for feature extraction. These datasets are explained briefly in the following sections:

\subsubsection{CASIA V2.0}
It contains 12,614 images in total, of which 7,497 are authentic, and 5,123 are forged images. The resolution of images ranges from 240 x 160 to 900 x 600. The tampered images have been applied to post-processing operations and saved in JPEG and TIFF formats. Out of these 5,123 tampered images, 3,274 images are copy-move, and 1,788 are splicing. The number of authentic images presents, respectively, for forged images, are 1,701. Henceforth, our total dataset size comes out to be of 4,975 images. 

\subsubsection{CoMoFoD}
This dataset contains 400 images, 200 authentic, and 200 forged. It contains only copy-move forgery cases in PNG format. The dimension of images in this dataset is 512 x 512. Various distortions such as translation, rotation, and scaling are applied to tampered images.

\subsection{Implementation Details}
We explored diverse color spaces to get a sense of the behavior of CMF images in different color spaces. Using Alexnet for feature extraction and DANN for domain adaptation, we varied the number of CMF images across RGB and YCrCb color space for the CASIA dataset. Chrominance component of YCrCb illuminates the identical regions in images with the same luminosity. It helps the deep networks to visualize copy-pasted regions in images.

In DANN, we used categorical cross-entropy as loss function and Adam optimizer with learning rate 0.001. The DDC network is trained using Stochastic Gradient Descent optimizer, with a momentum value of 0.9, and learning rate value of 0.0001. At the time of training, we initially used only CMF images for unsupervised domain adaptation. Then, we included semantic inpainted images to study the effects of edge discrepancy in recognizing forged images. There are no labels used at the time of training. For the target domain, images are passed with a domain label attached to it, and the source domain has a class label also assigned to it. The source model adapts the weights to classify target images with the same features into a particular category.

During testing, the target images are passed through the source classifier model, whose weights are now adapted to features specific to target data. We divided the CASIA dataset into an 80:20 ratio. 80\% of the data used for training, and then, 20\% used for testing. As CoMoFoD contains only 200 images, all the images were used to learn the discriminative features, as well as for evaluation. We used classification accuracy, precision, recall, and F1-score as performance metrics to evaluate our architectures. Precision is expressed as the number of true positives divided by the sum of true and false positives. The recall is defined as the ratio of true positives by true positives and false negatives. F1- score is the harmonic mean of recall and precision score. 

\subsection{Performance Analysis}

We will now discuss the results summarized in Table \ref{tab1}, \ref{tab2} and \ref{tab3}. We trained our architecture on source dataset and evaluated it on target dataset. From Table \ref{tab1}, we can visualize the performance of architecture in RGB and YCrCb space. 

\begin{table}[h!]
\begin{center}
\begin{tabular}[width=0.8\columnwidth]{|c|c|c|c|c|}
\hline
Number of & \multicolumn{2}{c|} {Accuracy Score}  & \multicolumn{2}{c|}{F1- Score} \\
\cline{2-5}
images used & RGB & YCrCb & RGB & YCrCb \\
\hline
20,000  & 59.76 & 57.62 & 72.12 &  70.47 \\
\hline
30,000 & 61.84 & 58.27 & 72.73 &  71.59\\
\hline
40,000 & 63.52 & 59.7 & 73.15 & 72.86 \\
\hline
50,000 & 63.91 & 62.78 & 76.51 & 73.02\\
\hline
60,000 & \textbf{65.39} & 64.29 & \textbf{78.83} & 77.97\\
\hline
\end{tabular}
\end{center}
\caption{Color space accuracy comparison with Copy-Move forged images only in RGB and YCrCb space}\label{tab1}
\end{table}

\begin{table*}[h!]
\begin{center}
\begin{tabular}[width=\textwidth]{|c|c|c|c|c|c|c|c|c|c|c|c|}
\hline
\multicolumn{4}{|c|}{Source Images} & \multicolumn{4}{c|}{COCO $\rightarrow$ CASIA} & \multicolumn{4}{c|}{COCO $\rightarrow$ CoMoFoD} \\
\cline{1-12}
\multicolumn{2}{|c|}{CMF} & \multicolumn{2}{c|}{Inpaint} & \multicolumn{2}{c|}{DDC} & \multicolumn{2}{c|}{DANN} & \multicolumn{2}{c|}{DDC} & \multicolumn{2}{c|}{DANN} \\
\cline{1-12}

Authentic & Forged & Authentic & Forged & Acc. & F1 & Acc. & F1 & Acc. & F1 & Acc. & F1 \\
\hline
10,000 & 10,000 & - & - & 60.63 & 71.92 & 59.76 & 71.47 & 60.25 & 61.52 & 51 & 63.7 \\
\hline
10,000 & 10,000 & 5,000 & 5,000 & 66.56 & \textbf{80.18} & 64.78 & 78.52 & \textbf{66.25} & \textbf{78.99} & 51.29 & 65.48 \\
\hline
10,000 & 10,000 & 10,000 & 10,000 & 64.26 & 77.74 & 64.85 & 76.54 & 63.25 & 76.12 & 51.71 & 63.38\\
\hline
15,000 & 15,000 & 5,000 & 5,000 & \textbf{67.97} & 76.62 & \textbf{65.91} & \textbf{79.16} & 57 & 76.05 & \textbf{51.75} & \textbf{67.12}\\
\hline
\end{tabular}
\end{center}
\caption{Evaluation of DANN and DDC trained with base feature extraction model as Alexnet varying the amount and type of source domain images.} \label{tab2}
\end{table*}

\begin{table*}[ht!]
\begin{center}
\begin{tabular}{|l|c|c|c|c|c|c|}
\hline
Architecture & \multicolumn{3}{c|}{CASIA} & \multicolumn{3}{c|}{CoMoFoD} \\
% Architecture & P CASIA & Recall & F1 & P CoMoFoD & Recall & F1 \\
\cline{2-7}
& Precision & Recall & F-1 & Precision & Recall & F-1 \\
\hline
\hline
BusterNet \cite{busternet} & \textbf{78.22} & 73.89 & 75.98 & 57.34 & 49.39 & 49.26\\
\hline
DDC MMD AlexNet & 68.02 & 97.73 & 80.18 & \textbf{65.62} & \textbf{97.86} & \textbf{78.99}\\
\hline
DDC MMD VGG & 57.27 & 95.02 & 70.13 & 65.39 & 91.24 & 76.18\\
\hline
DANN AlexNet & 66.19 & 98.47 & 79.16 & 51.05 & 83.5 & 63.34\\
% 50.76 & 99.01 & 67.12
\hline
DANN VGG & 68.26 & \textbf{99.84} & \textbf{80.34} & 49.65 & 72.04 & 58.75\\
\hline
\hline
\end{tabular}
\end{center}
\caption{Comparison from previous architecture BusterNet\cite{busternet}. Our model DANN with VGG feature extraction and DDC with AlexNet performs the best on CASIA and CoMoFoD dataset respectively.}\label{tab3}
\end{table*}

As the number of images increased, the results improved for domain adaptation. Due to complex post-processing operations, YCrCb space was unable to localize same tampered regions. As RGB color space performed better, therefore, for our future training of domain adaptation algorithms, we chose RGB images for source and target domains.

\begin{figure}[htbp]
\centering
\includegraphics[width=0.9\columnwidth]{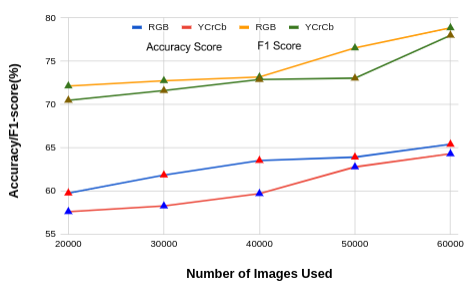} % Reduce the figure size so that it is slightly narrower than the column.
\caption{Accuracy/F1-score v/s Number of source images used. This plot shows the \textit{COCO $\rightarrow$ CASIA}  case.}
\label{fig5}
\end{figure}

In starting, we only used CMF images for unsupervised DA. DANN and DDC were able to minimize the distance between the two datasets distributions, but using only CMF images makes the network biased towards objects resembling the same feature characteristics. To analyze the contribution of the amount of CMF images for domain adaptation, we examined each time with an increment of 10,000 images. We saw that just by increasing CMF images, there are no noteworthy changes in the accuracy and F1-score on the target domain. In cases where the copied and background region are the same, \eg, grass, then the model is unable to distinguish the image as authentic or tampered. To alleviate this problem, we incorporated semantic inpainted images to learn the edge discriminative features. It helped the model to learn the dissimilarities near the edges of the images are copy-pasted. As the target domain contains CMF images, increasing the distribution of semantic images beyond 10,000 images leads to drop in performance. Table \ref{tab2} shows the effect of utilizing both semantic inpainted and copy-move tampered images. In contrast to contemporary networks such as Inception\cite{inception} and ResNet\cite{resnet}, we used AlexNet and VGG-7 as our base models, because, these networks have a huge number of parameters and due to limited amount of target domain images, the model doesn't generalize well.

\noindent \textbf{\textit{COCO $\rightarrow$ CASIA:}} In CASIA, there are 3979 images used for training purposes and 996 images for evaluation. With DDC, we saw a sudden jump by including semantic inpainted images. Using DANN, we achieved the best score, when the highest number of images were used. As there is a large number of images available at the time of training, we can see from Table \ref{tab3} that DANN+VGG-7  achieves the highest recall and F1-score. 

\noindent \textbf{\textit{COCO $\rightarrow$ CoMoFoD:}} CoMoFoD dataset is very small. Due to the presence of 200 images only, we trained and evaluated on the whole dataset. With the increase in the number of images in the source domain, accuracy and F1 score decreased in DDC, and, insignificant increase using DANN. As the dataset was small, we can see from Table \ref{tab3} that DDC+ MMD with Alexnet as base model performed better compared to VGG-7. VGG-7 has a huge number of parameters that can't be optimized; hence, they performed poorly at the test time. 

To compare with previous work, we analyzed our results with BusterNet architecture. They mainly took into account of CASIA CMF and CoMoFoD dataset. Other works, mainly used all images of CASIA dataset, not explicitly for CMF images. In BusterNet, they created and trained on 1 lakh images for supervised training, and then evaluated on these datasets. In CoMoFoD, they used 200 images as ours, but, in CASIA, they took only 1356 CMFD images into account compared to 4975 of ours. Our approach improves the accuracy by 5-6\% in the case of CASIA and 27-28\% in the case of CoMoFoD. Table \ref{tab3} shows the performance comparison between ours and BusterNet. Whereas BusterNet has used pixel-wise annotations to learn the class of images, we have not used any label at the time of training. In our case, as the data distribution is too much imbalanced, precision and recall score plays a significant role. We can see that our precision score is not in the comparable range of recall scores. It is due to the reason, as we have less number of positive class images in contrast to the negative class. As we look into the denominator of precision and recall, in the first case, the denominator is the sum of true plus false positives. Now, we have too many images in a false class. It attributes to a large number of false positives. Whereas in the recall, the denominator is the sum of true positives plus false negatives. The false-negative number is less as the number of images in the correct class is fewer.

\section{Conclusion}
In this paper, we suggested a new approach for data generation to counter the problem of small, publicly available datasets for image forgery. We outlined novel unsupervised learning approaches to detect CMF in images. We presented evaluation over different feature extraction models. Our approach outperforms the accuracy of the previous method, which incorporates supervised deep learning. In the future, we aim to explore GANs for the generation of realistic tampered images and the inclusion of splicing forgery to make our model more robust.

{\small
\bibliographystyle{ieee}
\bibliography{ref}
}

\end{document}